\documentclass{article}
\usepackage{spconf,amsmath,graphicx}
\usepackage{xcolor,graphicx,amssymb,multirow,multicol,booktabs}
\usepackage{svg}
\usepackage{pdfpages}
\usepackage{hyperref}

\graphicspath{ {./images/} }
\usepackage{stfloats}

\title{Scaling ASR Improves Zero and Few Shot Learning}
\name{\begin{tabular}{c} Alex Xiao*, Weiyi Zheng*, Gil Keren, Duc Le, Frank Zhang, Christian Fuegen\\Ozlem Kalinli, Yatharth Saraf, Abdelrahman Mohamed\thanks{*Equal contribution}\end{tabular}}
\address{Facebook AI}
\begin{document}
\raggedbottom
\ninept
\maketitle
\begin{abstract}
With 4.5 million hours of English speech from 10 different sources across 120 countries and models of up to 10 billion parameters, we explore the frontiers of scale for automatic speech recognition. We propose data selection techniques to efficiently scale training data to find the most valuable samples in massive datasets. To efficiently scale model sizes, we leverage various optimizations such as sparse transducer loss and model sharding. By training 1-10B parameter universal English ASR models, we push the limits of speech recognition performance across many domains. Furthermore, our models learn powerful speech representations with zero and few-shot capabilities on novel domains and styles of speech, exceeding previous results across multiple in-house and public benchmarks. For speakers with disorders due to brain damage, our best zero-shot and few-shot models achieve 22\% and 60\% relative improvement on the AphasiaBank test set, respectively, while realizing the best performance on public social media videos. Furthermore, the same universal model reaches equivalent performance with 500x less in-domain data on the SPGISpeech financial-domain dataset.

\end{abstract}
\section{Introduction}
\label{sec:intro}

Using massive datasets to train neural models with ever-increasing sizes has spurred rapid progress in many fields of machine learning, such as natural language processing \cite{gpt3,roberta}, computer vision \cite{goyal2021self}, and automatic speech recognition (ASR) \cite{speechstew,li2021scaling,zhang2020pushing}. 
The size of the training dataset and the number of model parameters are mutual bottlenecks and must be scaled in tandem \cite{scalinglaws}. In this paper, we explore and overcome the limitations of these two dimensions in ASR.

The abundance of publicly available text on the internet enabled the large-scale training of language representation models of up to 175B parameters on hundreds of billions of tokens \cite{gpt3}. On the other hand, supervised ASR datasets and models have been orders of magnitude smaller, and only recently, billion parameter ASR models are used with semi-/self-supervised methods \cite{zhang2020pushing,wav2vec,hsu2021hubert} or through pooling together data from many sources \cite{speechstew}.

This paper pushes these ideas to the extreme by pooling data from 10 different sources and employing semi-supervised training through pseudo-labeling. Our data contains 4.5 million hours of speech, most notably 4 million hours of unlabelled public social media videos on Facebook, uploaded from 120 countries and containing a wide variety of content and acoustic conditions. We propose data selection strategies to emphasize data diversity while reducing the computation cost of working with the whole dataset.

Following prior work on scaling Transformer models \cite{gpt3, vaswani2017attention,megatron}, we scale the encoder of an E2E VGG-transformer transducer model \cite{yeh2019transformer, wang2020transformer} up to 10B parameters. We leverage several techniques to train our transducer models efficiently on GPUs: FairScale model sharding \cite{FairScale2021}, sparse alignment restricted transducer loss \cite{mahadeokar2021alignment},  mixed-precision training \cite{micikevicius2017mixed}, and large batch sizes \cite{goyal2017accurate}.

\begin{table}
\centering
 \begin{tabular}{cccc}
  \toprule

\multirow{2}{*}{\textbf{Data Source}} & \multirow{2}{*}{\textbf{Transcriber}} & \multirow{2}{*}{\textbf{Hours}} & \textbf{Hours After}\\ && & \textbf{Augmentation} \\
 \midrule
 LibriSpeech \cite{panayotov2015librispeech} & Human & 960 & 5760 \\
 Common Voice \cite{ardila-etal-2020-common} & Human & 500 & 3000 \\
 Libri-Light \cite{kahn2020libri} & Model & 60000 & 360000 \\
 Fisher \cite{cieri2004fisher} & Human & 1960 & 11760 \\
 Assistant* & Human & 12600 & 41400 \\  % appen, hey fb
 Conversational* & Human & 780 & 6600 \\ % Appen data, viewpoints
 Calling Names  & TTS &640 & 3840 \\
 Dictation* & Model & 880 & 7920 \\ % Stella phase 2. can we talk about stella?
 Portal † & Human & 1350 & 8100 \\ 
 Video †  & Human & 18000 & 108000 \\
  Portal † & Model & 4800 & 28800 \\
 Video † & Model & 4009400 & 4009400 \\ 
 \bottomrule
 \end{tabular}
 \caption{Our 4.5M hour dataset consists of 10 sources. Data sources marked with * are collected through third-party vendors. Those marked with † are collected from Facebook products.}
  \label{tab1}
  \vspace{-10pt}
\end{table}

Prior work \cite{speechstew,zhang2020pushing,hsu2021hubert,robustw2v,RASR} explored mixing many datasets to train large multi-domain speech models but was limited to under 100K total hours and 1B model parameters. \cite{droppo2021scaling} analyzed scaling trends for acoustic models but did not go beyond 10K hours and 100M parameters. With a focus on multi-lingual models, \cite{li2021scaling} scaled ASR models up to 10B parameters but only demonstrated less than 0.5\% relative improvement compared to 1B parameter models. This paper expands these efforts to show that English speech has sufficient difficulty to merit scaling to 10B parameters and shares a recipe to train models at this scale efficiently.

While videos on social media are abundant, other scenarios severely lack audio resources. For example, AphasiaBank \cite{macwhinney2011aphasiabank}, the largest source for aphasic speech recognition, contains under 100 hours of audio data. By pushing the limits of scale for ASR, we can improve ASR not just for domains with large datasets but also low resource domains like aphasic speech. Pre-training large models on a universal dataset shows impressive zero-shot 22\% WER improvement on AphasiaBank. Transfer to other novel domains with zero, limited, and large-scale fine-tuning conditions exceed previously reported results, e.g., SPGISpeech \cite{o2021spgispeech} and an in-house dataset of long-form videos. We find scaling model size to 1B parameters to significantly improve zero and few-shot performance, even in low resource conditions.

\section{Data Scaling}

\subsection{Multi-domain Data Sources}
Our first method of constructing a large speech recognition dataset is to pool data from various sources. Table \ref{tab1} lists out the data sources used. The data can be grouped into four categories:
\begin{itemize}
\item Publicly released datasets: LibriSpeech \cite{panayotov2015librispeech}, Common Voice \cite{ardila-etal-2020-common}, Libri-Light \cite{kahn2020libri}, and Fisher \cite{cieri2004fisher}.
\item In-house datasets collected from third-party vendors via crowd-sourced volunteers responding to artificial prompts with mobile devices. The content varies from voice assistant commands to a simulation of conversations between people. 
\item In-house datasets from Facebook products: public Facebook videos and voice commands to Portal. Videos used are from 120 different countries. 
\item Data generated from an in-house TTS model to increase the diversity of sentence patterns in our training data. 
\end{itemize}

All in-house datasets are de-identified with no personally identifiable information (PII). Depending on the source, the data was further augmented with various distortion methods: speed perturbation \cite{ko2015audio}, simulated reverberation, and randomly sampled additive background noise extracted from public Facebook videos.

We retain punctuation and casing from in-house datasets, which introduces inconsistency with some public datasets but allows the final model to output richer information. For evaluation, we use hand-transcribed data from the LibriSpeech, Portal, Video, and Conversational data sources, ranging from 3K-15K utterances with no overlap with training. We split up Video into ``Standard" and ``Challenging" subsets, where the ``Challenging" subset contains videos with more noise and music.

\subsection{Semi-supervised Labeling}

The key to our data scaling strategy is leveraging 4 million hours of unlabelled audio with pseudo-labeling. The majority of our data in Table \ref{tab1} comes from public Facebook videos labeled by a 1B parameter model trained on a smaller supervised and semi-supervised dataset from similar sources. We use a language identification model to select videos predicted to be English. These videos came from the same source as the supervised videos but may contain more challenging data such as singing and foreign speech.

\subsection{Data Selection}

4M hours of pseudo-labels present many challenges, including noisy labels and audio, viral videos dominating most of the content, and infrastructure requirements to work with such a massive dataset. Data selection is a common technique to address these issues \cite{liao2013large}. We propose the following strategies for data selection to bring the dataset down to 1.3M hours only:

\begin{itemize}
    \item \textit{Words per Second}: Remove pseudo-labels with fewer than 0.5 words per seconds, which correlates with noisy music videos or foreign language.
    \item \textit{Confidence Score}: Remove data with a bottom 20\% confidence score to remove low confidence pseudo-labels.
    \item \textit{Model Disagreement}: Re-decode the unlabeled data with an 80M parameter streaming model. We compute the edit distance between the two hypotheses to filter out data within the bottom and top 20\% of disagreement to avoid too easy and too noisy utterances.
    \item \textit{Segmentation + Alignment}: A hybrid model \cite{chenones} is used for the alignment restricted loss \cite{mahadeokar2021alignment} to segment data into 10s segments and filter out empty segments or ones that fail to align.
    \item \textit{Rare Data}: Compute the cumulative word frequency distribution based on the supervised data and a consider a word to be rare if it is not in the top 90\% most frequent words. Video segments with $W$ words and $R$ rare words are preserved if $R >= min(2, 0.25 * W)$. We keep all videos not uploaded from United States, Great Britain, Canada, or Australia to maintain data diversity. %(91\% of videos come from these countries).
\end{itemize}

\begin{table}
\centering
 \begin{tabular}{cccc}
  \toprule
\textbf{Parameters} &\textbf{Hidden Size} &\textbf{Layers} & \textbf{Attention Heads} \\
 \midrule
 100M & 512 & 36 & 8\\
 1B & 1152 & 60 & 16 \\
 10B & 3072 & 90 & 48 \\
 \bottomrule
 \end{tabular}
 \caption{Hyper-parameters for our Transformer encoders.}
  \label{tab2}
  \vspace{-10pt}
\end{table}

\section{Model Scaling}
\label{sec:model}

\subsection{Model Architecture}
Our model architecture is a non-streamable full-context Transformer-Transducer \cite{yeh2019transformer} with a VGG-Transformer encoder \cite{wang2020transformer}, a 19M parameter 2-layer LSTM predictor, and a 4M parameter feed-forward joiner. We focus on increasing the size of the Transformer encoder, which showed the most promise in initial experiments. Three encoders of 100M, 1B, and 10B parameters are constructed by varying the number of transformer layers and hidden dimensions. FFN dimension is always set to 4 times the hidden dimension. 3 VGG blocks are applied at the encoder input \cite{wang2020transformer} for an inter-frame length of 80ms. We use 0.1 dropout in all Transformer blocks. Details for each model size are listed in Table \ref{tab2}. We also experimented with Mixture of Experts encoders \cite{fedus2021switch} of up to 40B parameters. We did not see improvements, hence, we leave its exploration for future work.

\begin{figure*}[t]
\centering
\vspace{-0.2cm} % Seems a little empty without this
\includegraphics[width=\textwidth]{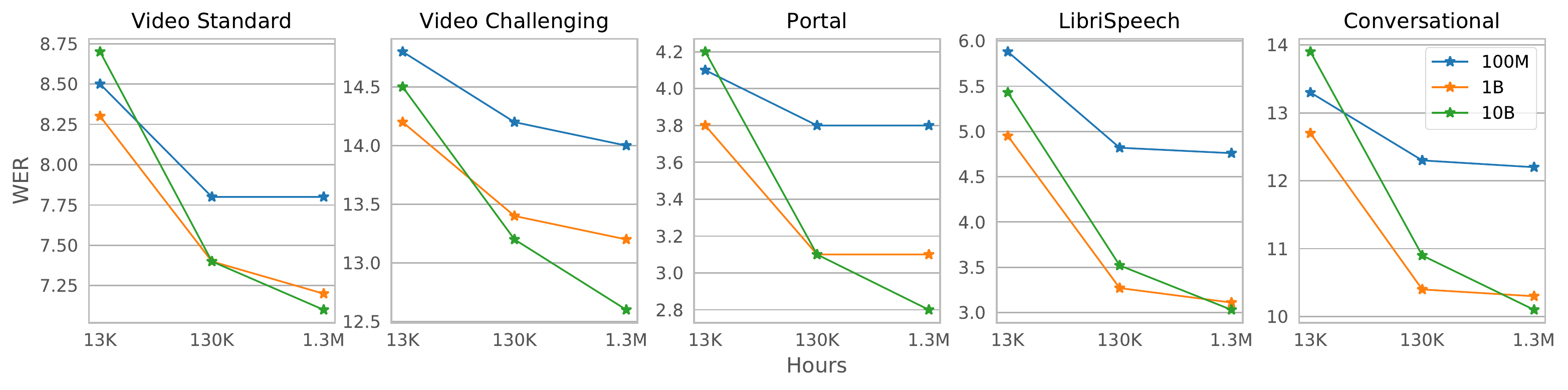}

\vspace{-0.25cm} % Seems a little empty without this
\caption{WER results of a single universal model as we vary model size (100M, 1B, and 10B parameters) and dataset size (13K, 130K, and 1.3M hours). LibriSpeech results are reported on the test-other set. Increasing model size from 1B to 10B parameters only helps in the largest data setting: we see an average relative WER change of 7.32\%, 2.19\%, and -4.03\% on 13K, 130K, and 1.3M hours respectively.}
\label{fig1}
\vspace{-10pt}
\end{figure*}
\subsection{Model Convergence}
Due to convergence stability challenges in large mixed-precision models \cite{bachlechner2020rezero} with gradients or activations overflowing we recommend the following strategies:

\begin{itemize}
    \item Pre-layernorm \cite{xiong2020layer} avoids gradient explosion and enables better gradient flow.
    \item Scale the weight of the second linear layer in the FFN block by $\frac{1}{\sqrt{2n}}$, where $n$ is the number of Transformer blocks \cite{megatron}. 
    \item Set $\beta_2$ in the Adam optimizer to $0.98$ to avoid network activations overflowing beyond the FP16's range \cite{roberta}.
\end{itemize}

\subsection{Model Training Efficiency}
Training transducer models with a billion or more parameters with distributed data parallel training  is prohibitively slow. We leverage multiple optimizations to make it feasible to train such models in a reasonable amount of time. Large batch sizes can speed up training by improving the efficiency of GPU kernels and reducing the number of inter-GPU communication rounds required \cite{goyal2017accurate}. We use a global batch size of 23 hours. 
To fit such a large batch size into GPU memory, we leverage the following optimizations: 
\begin{itemize}

    \item \textit{Alignment Restricted Transducer Loss} \cite{mahadeokar2021alignment} utilizes word level alignments to reduce the memory required for transducer loss from $O(B \times T \times U \times D)$ to $O( B \times (T + U \times (b_l + b_r)  )\times D)$, where $B$ is the batch size, $T$ the number of timesteps, $U$ the number of target symbols, $D$ the vocabulary size, and $b_l$ and $b_r$ are the left and right buffers. We set $b_l = 15$ and $b_r = 15$. 
 
    \item \textit{Fully Sharded Data-Parallel} \cite{FairScale2021} shards model weights, gradients, and optimizer states to reduce the memory consumption of large models. We only shard optimizer state and gradients   to reduce the communication overhead.
    \item \textit{Activation Checkpointing} \cite{chen2016training} reduces activation memory by recomputing them in the backward pass. 
    \item \textit{Mixed Precision Training} \cite{micikevicius2017mixed} utilizes GPU Tensor Cores for more efficient compute and reduces the GPU memory and communication bandwidth required. 
\end{itemize}

\section{Experiments}
\subsection{Experiment Details}

We use 80-D log Mel features computed every 10ms with a window of 25ms. SpecAugment \cite{park2019specaugment} with the LibriSpeech Double policy is applied to the input features. We train our models for 200,000 updates, linearly increasing the learning rate to $4e^{-4}$ in the first 20,000 updates and exponentially decaying by $1e^{-2}$ over the remaining updates. We use Adam with $\beta_1 = 0.9, \beta_2 = 0.98, \epsilon=1e^{-6}$ and normalize the global gradient norm to 2. The vocabulary is set to 4095 BPE units. All training is done in Fairseq \cite{ott2019fairseq}. 
Our largest 10B parameter model is trained with 128 A100 GPUs for 25 days and needs $8.41 * 10^{6}$ PFLOPs for the encoder.

\subsection{Impact of Scaling}

To analyze the interaction between data and model size, we train a universal model with three different sizes on three datasets of 13K hours, 130K hours, and 1.3M hours. Results are plotted in Figure \ref{fig1}. 
We find that there is a benefit when scaling dataset size and model size together. At 1.3M hours, WER reduction is correlated with the model size where the 10B model obtains on average a 4.03\% relative WER reduction compared to the 1B model and 20.00\% relative reduction compared to the 100M model. 
Similarly, WER reduction is correlated with dataset size at 10B parameters. Increasing 130K hours to 1.3M hours improves the average relative WER of the 100M model by 0.01\%, the 1B model by 1.67\%, and the 10B model by 8.46\%. These results suggest that scaling the model and dataset together is the key to further improvement.

\begin{table}
\centering
 \begin{tabular}{cccc}
  \toprule
{\textbf{Model Size}} & {\textbf{Data Size (h)}} & {\textbf{WER}} & \textbf{Rare WER}\\ 
 \midrule
  10B & 3.2M  & 7.21 & 11.00 \\
 10B  & 1.3M & \textbf{7.16} & \textbf{10.55} \\
  1B & 3.2M  & 7.56 & 11.53 \\
 1B & 1.3M  & \textbf{7.46} & \textbf{10.88} \\
 \bottomrule
 \end{tabular}
 \caption{Effect of data selection with \textit{Rare Data} and \textit{Model Disagreement} after 100k updates on average WER and Rare WER.}
  \label{tab3}
  \vspace{-10pt}
\end{table}

\subsection{Data Selection}
Applying all data selection methods described in Section 2 reduced the original data from 4.5M hours to 1.3M hours. Without applying \textit{Rare Data} and \textit{Model Disagreement} filtering on the pseudo-labels, the dataset is about 3.2M hours. The goal of these two techniques is to reduce cost by removing unnecessary data while improving performance on the long tail. To measure the impact of these two methods, we introduce the rare WER metric, which measures WER only on words outside the top 90\% cumulative word frequency distribution -- computed on the supervised data. These words are often proper nouns and more important to the meaning of the utterance than common words like articles. Table \ref{tab3} shows that reducing the data by 1.9M hours not only maintains the overall WER but also improves rare WER by 4-6\% relative. Although we previously found increasing dataset size beneficial, these findings suggest that quality is more important than quantity: it is crucial to pick diverse samples when scaling up dataset size.

\begin{table*}

\centering
 \begin{tabular}{llcccccccccc}
  \toprule
 \textbf{Dataset} && \multicolumn{2}{c}{\textbf{AphasiaBank}}   &\multicolumn{2}{c}{\textbf{Long-Form Video}} & \multicolumn{5}{c}{\textbf{SPGISpeech}} \\ 
 
 \cmidrule(lr){3-4} \cmidrule(lr){5-6}  \cmidrule(lr){7-11}
 
&& \textbf{Overall} &  \textbf{Fold 1} & \textbf{Short} & \textbf{Long} & \textbf{5000h} & \textbf{100h} & \textbf{10h} & \textbf{1h} & \textbf{10m}  \\

 \toprule
 
\textbf{Prior Work} & & 37.37 \cite{macwhinney2011aphasiabank} & - & - & - & 2.3\textsuperscript{*} \cite{o2021spgispeech} & -  & - & - & - \\
  \midrule

\multirow{3}{*}{\textbf{From Scratch}} & 100M & \textbf{53.39} & \textbf{51.72} & \textbf{13.52} & \textbf{8.94} & 2.6(2.5) & 18.5(18.5) & - & - & - \\

&1B & 54.32 & 52.51 & 13.56 & 9.01 & 2.6(2.5) & \textbf{17.1}(17.0) & - & - & - \\

&10B& 56.69 & 54.81 & 15.30  & 10.14 & \textbf{2.4}(2.4) & 27.9(27.8) & - & - & - \\

 \midrule

\multirow{3}{*}{\textbf{Universal}} & 100M & 30.29 & 29.63 & 12.99 & 9.17 & \multicolumn{5}{c}{7.1(4.9)}\\
& 1B & \textbf{29.06} & \textbf{28.55} & 11.88 & \textbf{8.72} & \multicolumn{5}{c}{6.5(4.4)} \\

& 10B & 30.05 & 29.33 & \textbf{11.43} & 9.33 &   \multicolumn{5}{c}{\textbf{6.4}(4.3)}  \\
 
 \midrule
 
\multirow{3}{*}{\textbf{+ Fine-tuning}} & 100M & 16.44 & 15.59 & 12.68 & 8.45 & 2.0(2.0) & 2.7(2.6) & 3.0(2.9) & 3.5(3.4) & 4.2(3.9) \\

&1B & \textbf{14.83} & \textbf{13.98} & 11.18 & \textbf{7.45} & \textbf{1.8}(1.8) & \textbf{2.2}(2.2) & \textbf{2.4}(2.3) & \textbf{2.7}(2.6) & \textbf{3.9}(3.4)\\

&10B & 15.76 & 15.11 & \textbf{11.09} & 8.21 & \textbf{1.8}(1.7) & \textbf{2.2}(2.1) & \textbf{2.4}(2.4) & 2.9(2.8) & 4.0(3.4) \\

 \bottomrule
 \end{tabular}
 \caption{Results on novel domains. We benchmark 3 types of models: universal (trained on the general 4.5M hour dataset), from scratch (trained on the in-domain dataset), and fine-tuned (fine-tune the universal model on the in-domain dataset). Fine-tuning is significantly better than training from scratch and enables 1B+ models on lower resource domains. We also report WER computed with an in-house reference normalization in parentheses. \textsuperscript{*}Private test we don't have access to.} 
  \label{tab4}
  \vspace{-10pt}
\end{table*}

\subsection{Zero-shot and Few-shot ASR}

To understand how our models generalize to novel domains, we perform zero-shot and few-shot experiments on three datasets:  AphasiaBank \cite{macwhinney2011aphasiabank}, SPGISpeech \cite{o2021spgispeech}, and an in-house long-form videos dataset. We conduct few-shot learning by fine-tuning the universal models from Figure \ref{fig1} further on each respective dataset. Our models achieve strong zero-shot performance and demonstrate impressive few-shot performance by exceeding baseline results by 16\% to 60\% relative (Table \ref{tab4}). In all cases, few-shot learning on top of our universal model is significantly superior to training on the relevant domain from scratch, enabling low-resource domains to enjoy the benefits of large models.

Our experiments also show that zero-shot and few-shot learning benefit from scaling from 100M to 1B parameters. The 10B results, however, are less consistent and points to overfitting during fine-tuning. We highlight our results below.

\subsubsection{AphasiaBank}
Aphasia is an acquired speech-language disorder due to damages to portions of the brain, most commonly resulting from a stroke. It impairs verbal communication and makes it difficult for ASR systems to understand aphasic speech \cite{Le2016aphasiabank,Le2018quantitative}. Transcribed aphasic speech is also scarce: a large-scale aphasic speech dataset like AphasiaBank \cite{macwhinney2011aphasiabank} only contains about 100 hours of recorded interactions between clinicians and persons with aphasia (PWAs). These challenges motivate leveraging transfer learning from a large, diverse dataset like ours. We hope that high-quality ASR for aphasic speech will allow PWAs to enjoy the benefits of ASR technologies while enabling medical analyses that rely on ASR \cite{Le2018quantitative}.
 
We follow the same normalization and data folds from \cite{Le2016aphasiabank}. Results aggregated across all four folds are shown in Table \ref{tab4}. When trained from scratch, large E2E models cannot achieve WER better than 50\%. On the other hand, universal and fine-tuned models perform quite well; the fine-tuned 1B parameter model achieves a 60\% relative WER improvement compared to the baseline hybrid model in \cite{Le2018quantitative} and a 72\% relative WER improved compared to our own baseline, both of which were trained from scratch on AphasiaBank. These results indicate that few-shot learning benefits low resource domains like aphasic speech. More work needs to be done, however, to avoid overfitting for the 10B model. We also note that concurrent work \cite{torre2021improving} achieved a WER of 28.5 on AphasiaBank, but the difference in data splits prevents a formal comparison.

\subsubsection{SPGISpeech}
SPGISpeech \cite{o2021spgispeech} contains 5,000 hours of financial audio from corporate earnings calls. We use the \texttt{norm} setting to analyze generalization to a more formal setting with financial jargon. The test set is private, so we split half of the 100h validation to create our own test set.

Table \ref{tab4} demonstrates that universal models perform somewhat reasonably but still struggle relative to \cite{o2021spgispeech}. Many errors are from jargon like ``GAAP" or mismatch in transcription conventions: ``uh" and ``um" insertions make up 1/4 of the errors. With our in-house reference normalization which avoids counting fillers as errors, the WER drops by about 30\% relative. After fine-tuning, our 10B model without normalization improves by 23\% relative compared to \cite{o2021spgispeech}.

We create smaller training sets with as little as 10 minutes of data to stress test low-resource adaptation. Our models display powerful adaptation capabilities: only 10 hours of fine-tuning data is needed to match the training performance from scratch on the original 500x larger dataset. Furthermore, few-shot learning improves WER by 20\% relative using only 10 minutes of data.  The 1B model performs the best while the 10B model overfits in ultra low-resource conditions. In contrast, the 1B model's extra capacity improves generalization from 5K hours to 1 hour; the improvement relative to 100M parameters steadily increases from 13\% to 23\%, suggesting a sweet spot for fine-tuning towards low resource domains.

\subsubsection{Long-Form Video}
We use 18000 hours of human-labeled long-form videos from social media to test the ability of our models to generalize to different lengths. These videos were in the original 4.5M hour dataset but with different lengths. We segment the training data to 45s instead of 10s and do not segment evaluation data. The evaluation videos are at most 5 minutes in length. Table \ref{tab4} breaks down the results into short videos (less than 45s) and long videos (more than 45s).

Within the universal models, the 10B model does the best on short videos but the worst on long videos, which suggests that although our huge dataset may be diverse in some areas, length diversity is still a blind spot for the large models. Fine-tuning alleviates this problem, but the 1B model still has a 9\% lower WER on long videos. This observation highlights the need for including length diversity when building large-scale datasets or regularization techniques to avoid overfitting to specific lengths.

\section{Conclusion}
In this work, we pushed the boundaries of large-scale speech recognition. We proposed an efficient recipe to train models of up to 10B parameters on 4.5M hours of audio. These large models demonstrated powerful zero-shot and few-shot learning capabilities across several domains, even with limited in-domain data. We also identified issues related to generalization and over-fitting in our current paradigm for scaling to 10B parameters. For future work, we plan to explore better low-resource transfer learning techniques for huge models. We will also investigate ways to improve data diversity and training objectives when working with massive datasets.

\newpage
\footnotesize
\bibliographystyle{IEEEbib}
\bibliography{refs}

\end{document}